\documentclass{wscpaperproc}
\usepackage{latexsym}
\usepackage{graphicx}
\usepackage{mathptmx}
\usepackage[T1]{fontenc}

\usepackage{amsmath}
\usepackage{amsfonts}
\usepackage{amssymb}
\usepackage{amsbsy}
\usepackage{amsthm}

\usepackage{soul}	
\usepackage{multirow}

\usepackage[noend]{algpseudocode}
\usepackage{algorithm}

\usepackage[pdftex,colorlinks=true,urlcolor=blue,citecolor=black,anchorcolor=black,linkcolor=black]{hyperref}

\newtheoremstyle{wsc}
{3pt}
{3pt}
{}
{}
{\bf}
{}
{.5em}
{}

\theoremstyle{wsc}

\begin{document}

%
%

\pagestyle{fancyplain}

\thispagestyle{plain}
\firstPageHead{}

\chead{\fancyplain{}{\itshape Giabbanelli and Beerman}}

\rhead{}
\cfoot{}
\renewcommand{\headrulewidth}{0pt} 

\makeatletter
\let\@internalcite\cite
\def\cite{\def\@citeseppen{-1000}%
    \def\@cite##1##2{(##1\if@tempswa , ##2\fi)}%
    \def\citeauthoryear##1##2##3{##1 ##3}\@internalcite}
\def\citeNP{\def\@citeseppen{-1000}%
    \def\@cite##1##2{##1\if@tempswa , ##2\fi}%
    \def\citeauthoryear##1##2##3{##1 ##3}\@internalcite}
\def\citeN{\def\@citeseppen{-1000}%
    \def\@cite##1##2{##1\if@tempswa, ##2)\else{}\fi}%
    \def\citeauthoryear##1##2##3{##1 (##3)}\@citedata}
\def\citeA{\def\@citeseppen{-1000}%
    \def\@cite##1##2{(##1\if@tempswa , ##2\fi)}%
    \def\citeauthoryear##1##2##3{##1}\@internalcite}
\def\citeANP{\def\@citeseppen{-1000}%
    \def\@cite##1##2{##1\if@tempswa , ##2\fi}%
    \def\citeauthoryear##1##2##3{##1}\@internalcite}
\def\shortcite{\def\@citeseppen{-1000}%
    \def\@cite##1##2{(##1\if@tempswa , ##2\fi)}%
    \def\citeauthoryear##1##2##3{##2 ##3}\@internalcite}
\def\shortciteNP{\def\@citeseppen{-1000}%
    \def\@cite##1##2{##1\if@tempswa , ##2\fi}%
    \def\citeauthoryear##1##2##3{##2 ##3}\@internalcite}
\def\shortciteN{\def\@citeseppen{-1000}%
    \def\@cite##1##2{##1\if@tempswa, ##2\else{}\fi}%
    \def\citeauthoryear##1##2##3{##2 (##3)}\@citedata}
\def\shortciteA{\def\@citeseppen{-1000}%
    \def\@cite##1##2{(##1\if@tempswa , ##2\fi)}%
    \def\citeauthoryear##1##2##3{##2}\@internalcite}
\def\shortciteANP{\def\@citeseppen{-1000}%
    \def\@cite##1##2{##1\if@tempswa , ##2\fi}%
    \def\citeauthoryear##1##2##3{##2}\@internalcite}
\def\citeyear{\def\@citeseppen{-1000}%
    \def\@cite##1##2{(##1\if@tempswa , ##2\fi)}%
    \def\citeauthoryear##1##2##3{##3}\@citedata}
\def\citeyearNP{\def\@citeseppen{-1000}%
    \def\@cite##1##2{##1\if@tempswa , ##2\fi}%
    \def\citeauthoryear##1##2##3{##3}\@citedata}
%
%
%
\def\@citedata{%
    \@ifnextchar [{\@tempswatrue\@citedatax}%
                  {\@tempswafalse\@citedatax[]}%
}

\def\@citedatax[#1]#2{%
\if@filesw\immediate\write\@auxout{\string\citation{#2}}\fi%
  \def\@citea{}\@cite{\@for\@citeb:=#2\do%
    {\@citea\def\@citea{, }\@ifundefined
       {b@\@citeb}{{\bf ?}%
       \@warning{Citation `\@citeb' on page \thepage \space undefined}}%
{\csname b@\@citeb\endcsname}}}{#1}}%

%
\def\@citex[#1]#2{%
\if@filesw\immediate\write\@auxout{\string\citation{#2}}\fi%
  \def\@citea{}\@cite{\@for\@citeb:=#2\do%
    {\@citea\def\@citea{; }\@ifundefined
       {b@\@citeb}{{\bf ?}%
       \@warning{Citation `\@citeb' on page \thepage \space undefined}}%
{\csname b@\@citeb\endcsname}}}{#1}}%

%
\def\@biblabel#1{}
\makeatother



\newdimen\bibindent
\bibindent=0.0em
\def\thebibliography#1{\section*{\refname}\list
   {}{\settowidth\labelwidth{[#1]}
   \leftmargin\parindent
   \itemindent -\parindent
   \listparindent \itemindent
   \itemsep 0pt
   \parsep 0pt}
   \def\newblock{}
   \sloppy
   \sfcode`\.=1000\relax}


\setlength{\baselineskip}{12.7pt}

\title{ACCELERATING HYBRID AGENT-BASED MODELS AND FUZZY COGNITIVE MAPS:\\HOW TO COMBINE AGENTS WHO THINK ALIKE?}

\author{\begin{center}Philippe J. Giabbanelli\textsuperscript{1} and Jack T. Beerman\textsuperscript{2}\\
[11pt]
\textsuperscript{1}Virginia Modeling, Analysis and Simulation Center, Old Dominion University, Norfolk, VA, USA\\
\textsuperscript{2}School of Data Science, University of Virginia, Charlottesville, VA, USA\end{center}
}

\maketitle

\vspace{-12pt}

\section*{ABSTRACT}
While Agent-Based Models can create detailed artificial societies based on individual differences and local context, they can be computationally intensive. Modelers may offset these costs through a parsimonious use of the model, for example by using smaller population sizes (which limits analyses in sub-populations), running fewer what-if scenarios, or accepting more uncertainty by performing fewer simulations. Alternatively, researchers may accelerate simulations via hardware solutions (e.g., GPU parallelism) or approximation approaches that operate a tradeoff between accuracy and compute time. In this paper, we present an approximation that combines agents who `think alike', thus reducing the population size and the compute time. Our innovation relies on representing agent behaviors as networks of rules (Fuzzy Cognitive Maps) and empirically evaluating different measures of distance between these networks. Then, we form groups of think-alike agents via community detection and simplify them to a representative agent. Case studies show that our simplifications remain accuracy.

\section{INTRODUCTION}
\label{sec:intro}

Agent-Based Modeling (ABM) serves to represent physical entities as virtual entities in a simulation for a plethora of real world scenarios. These entities known as `agents' can imitate living organisms such as cells or physical objects such as cities. Furthermore, ABMs are constructed with specific rule sets and laws governing the actions of agents depending on the problem that modelers are attempting to analyze. Regardless of the problem at hand, there is a balance to be struck between the size of the agent population and the resources required for computation. While improving model resolution (i.e., a simulated agent represents fewer real-world individuals) allows for detailed analyses in sub-groups, we must find a compromise given the limited resources such as compute time. This can be achieved through parallelism~\cite{buabeanu2023adaptive} and/or hardware accelerators such as GPUs~\shortcite{lysenko2008framework,Ghumrawi} or FPGA~\shortcite{xiao2019survey}. However, modelers may not possess such resources or the programming skills to efficiently use them. An alternative is to group agents. Such groups can be achieved by \textit{locally representative agents}, also known as `super agents' or `super nodes'~\cite{lippe2019using,parry2008comparative,parry2012large}. Agents can also be aggregated/disaggregated via a higher level of abstraction such as a continuum model~\cite{cilfone2015strategies,minucci2024agent}. In this paper, we focus on creating super agents to reduce the size of a model and accelerate simulations.

\begin{figure}[ht!]
    \centering
    \includegraphics[width=0.9\textwidth]{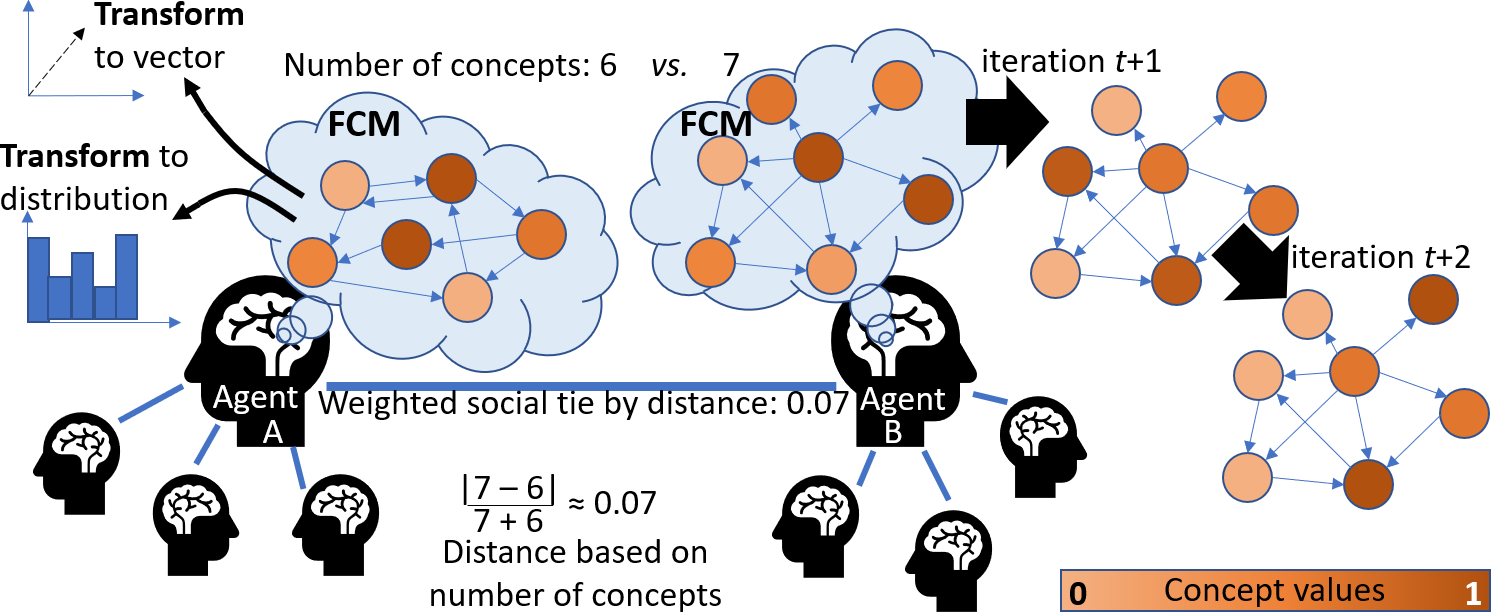}
    \caption{A hybrid ABM/FCM consists of agents who interact with each other (e.g., $A$ interacts with $B$ and each one has three other neighbors). Interactions are impacted by their `mental model' in the form of an FCM, which is a network that performs simulations. As exemplified for agent $B$, simulating an FCM changes its node values (in the interval [0, 1]) over discrete iterations. To compare mental models, we can compare FCMs by transforming these networks into distributions (e.g., degree distribution) or vector embeddings, or on the basis of simple criteria such as the number of nodes. Each existing social tie is then weighted to reflect the similarity of mental models. Here, similarity in number of nodes is about 0.07.}
    \label{fig:FCM}
\end{figure}

In order to find an agent whose behavior is representative of its group, we need to find a meaningful group (e.g., via community detection) and \textit{compare behaviors}. When the behavior of an agent is primarily a function of its individual traits (e.g., agents have different age or socio-economic status), the comparison is based on the distance between the vectors of traits, as exemplified by the Axelrod model of culture. However, real-world individuals with the same traits can still engage in different behaviors. This is not due to randomness, but to the fact that \textit{people follow different rules}: even if two seemingly indistinguishable individuals are presented with the same evidence, they may reach different conclusions. We thus use a hybrid modeling technique that combines ABMs with Fuzzy Cognitive Maps (FCMs)~\cite{davis2020fuzzy}. Intuitively, each agent has its own rule set (as shown with two agents in Figure~\ref{fig:FCM}), which is a simulation model consisting of a directed network with labeled nodes and weighted, directed edges\footnote{Node values range from 0 (absence of a concept) to 1 (full presence of a concept). Edge values range from -1 (an increase in the source \textit{decreases} the target) to 1 (an increase in the source \textit{increases} the target). An agent's view of causation is held constant hence edge weights do not change. Observations and circumstances change, hence node values are updated.}. After an agent interacts with peers or the environment, the observations provide an input to this FCM (similar to the `virtual brain' of the agent), which iteratively updates the node weights until reaching stabilization for the next decision (as shown for agent $B$ with two iterations in Figure~\ref{fig:FCM}). This hybrid method allows to represent the heterogeneity of human behaviors. It also illustrates the importance of a compromise since it carries a significant computational cost and even hardware accelerators are limited to populations of dozens of thousands of agents~\shortcite{Ghumrawi}.

Our main contribution is to reduce the population sizes of a hybrid ABM/FCM model by establishing representative agents within their communities while maintaining accuracy. This is achieved by implementing new and existing techniques to compare the mental models of agents, followed by community detection and aggregation. Ultimately our work contributes in two areas: \textit{(i)} we reduce computational costs of existing hybrid models, and \textit{(ii)} we introduce and evaluate new metrics to compare the behavior of agents.

Since behaviors are encoded through a network representation, we compare the behaviors of agents by comparing their FCMs. This becomes a matter of network differences, which can be expressed through network-level metrics (e.g., number of nodes) or by comparing vector- or distribution-based representation of the network (Figure~\ref{fig:FCM}). To keep the paper self-contained, we summarize classic network measures in Section~\ref{secGED:Backgroundmetrics} along with several new proposed measures. We also provide a succinct introduction to the hybrid ABM/FCM paradigm, which has been covered in more details at WinterSim previously. In Section~\ref{secGED:methodsmetrics} we describe the complete process to compare agents, starting with weighing social ties between agents represent the similarity/dissimilarity between their FCMs, and then running community detection algorithms to find clusters of behaviors. Agents are selected from these clusters as representatives and reconnected to form smaller hybrid models that are simulated again. In Section~\ref{secGED:Resultsmetrics}, we compare the simulations from the smaller model with the original one, in order to assess the effect of simplification onto simulation fidelity. 

\section{BACKGROUD}
\subsection{A Brief Introduction to Hybrid Agent-Based Models and Fuzzy Cognitive Maps}
The heterogeneity of real-world individuals stems not only from looking different or living in different contexts, but also because people \textit{reflect} differently on facts and experiences. In their reflections, people may be vague or uncertain, leading to fuzzy notions (``if it's too sunny I'll put on some sunscreen'') rather than crisp rules (``If UV Index Scale > 4 then apply 30 mg of SPF50 sunscreen''). As they make decisions, individuals can consider a variety of pros and cons, which can be interrelated or form patterns such as cycles. For instance, an individual whose family is facing a financial abyss may consider taking their own lives, which would result in even greater financial insecurity (and trauma) for their family. Fuzzy Cognitive Maps (FCMs) allow to transform the implicit mental models of individuals into explicit simulation models where factors (nodes) can be connected (edges), and these causal connections have numerical weights obtained via fuzzy logic. FCMs have been used in over 20,000 studies, often to model the perspectives of stakeholders in complex socio-environmental problems~\shortcite{Giabbanelli2024}. FCMs have also been integrated to ABMs, where each agent can be equipped with an FCM that has a unique structure and unique concept values, resulting in heterogeneity of behaviors~\cite{Giabbanelli2024b}. Computationally, an FCM is akin to a neural network, hence it is defined by (i) the network \textit{structure} of concepts and directed, weighted causal edges, representing an agent's ruleset; (ii) the current values of the agent's concept, known as \textit{activation vector}; and (iii) a \textit{transfer function}. In the same way as a neural network, a simulation is performed by applying the transfer function repeatedly to update the concept values, until either reaching stabilization or exceeding a user-defined number of iterations. 

\subsection{Comparing Behaviors via Networks: New and Established Measures}
\label{secGED:Backgroundmetrics}

We can implement a variety of metrics to characterize the Fuzzy Cognitive Map of an agent~\cite{gray2019assessing} and hence compare maps~\cite{tchupo2022comparing,wills2020metrics}. However, using the most appropriate metric is currently an open problem, hence this paper covers many alternatives to evaluate their impact empirically. Since the behavior of each agent is defined by their FCM, we consider metrics that are applicable to FCMs as well as metrics used for networks more broadly. The list of metrics is summarized in Table~\ref{tab:metricsinnovation} along with their equation, while this section summarizes the logic of each metric, including established metrics in network science (to keep this paper self-contained). We start with simple and quickly computed metrics (density, number of concepts, receiver/transmitter ratio, clustering coefficient) and then present more computationally intensive measures.

\textbf{Graph Density} is the number of connections in a network with respect to the maximum number of connections possible for all concepts. It depicts the interwovenness of concepts within an FCM~\cite{tchupo2022comparing}. The density of a network is calculated depending on whether it is directed (as shown in Table~\ref{tab:metricsinnovation}) or not. The \textbf{number of concepts} simply compares the total number of nodes in a network to the total number of nodes in the other network~\cite{tchupo2022comparing}. Intuitively, it can capture that the behavior of one agent depends on more concepts than another agent. The \textbf{receiver-transmitter ratio} is computed based on the number of `receiver nodes' (denoted as $R$), which serve exclusively as targets and have no outgoing edge, and the number of `transmitter' nodes (denoted as $T$) that always act as a source and have no incoming edge~\cite{tchupo2022comparing}. The \textbf{clustering coefficient} measures the average density of a node's neighbors. That is, we measure the coefficient for each node by looking at the density in the subgraph limited to the node's neighbors. Then, we average the coefficient across the nodes in order to get a clustering coefficient at the level of the graph. The clustering coefficient of a node is given in Table~\ref{tab:metricsinnovation} for the directed case. A node without neighborhood is a special case with a clustering of 0.

The measures above can extract one number for each FCM (e.g., number of nodes, graph density) thus two FCMs would be compared through two numbers. However, this may oversimplify important patterns. Instead of summarizing an FCM to a single number, we can thus extract more characteristics and convey them either through a vector or as a discrete distribution.

\textbf{Weighted Jaccard similarity} measures the distance between vectors, where all entries $x_i$ and $y_i$ are positive real numbers. First the similarity coefficient is determined through the summation of the minimum $x_i$ and $y_i$ values divided by the summation of the maximum $x_i$ and $y_i$ values. The \textit{distance} is the inverse of the similarity. In network science, the Jaccard similarity is often defined between two nodes on the basis of the intersection and union of their neighborhood sets~\cite{9820701}. However, we use the Jaccard similarity between discrete distributions (e.g., the degree distribution) at the network-level, rather than between sets of neighbors at the node-level. For example, our measure takes two FCMs' set of weighted edges and computes the Jaccard similarity coefficient. This allows us to compute a distance between the rulesets of agents.

The difference between two rule sets can be measured based on what is most `important' to each agent. The importance of a node is its  \textbf{centrality}, which produces a ranking from most to least important nodes. Common centrality measures include betweenness, closeness, and degree. To compare the rankings of two agents' FCMs, we consider rankings as vectors and perform a cosine similarity. The output ranges from -1 (least similar) to 1 (most similar)~\cite{7577578}. 

\textbf{Graph kernels} decompose a network into subtructures, such as a comprehensive inventory of all trees or loops. A common method is the Triad Significance Profile (TSP) that extracts all 16 possible subgraphs with three nodes (i.e., triads)~\cite{milo2004superfamilies}. Their significance is assessed statistically via the Z-score, denoted $Z_{M}$, where $n_{M}$ represents the frequency of a triad $M$ in the given network, and $\left\langle {n}_M^{\mathrm{rand}}\right\rangle$ and ${\sigma}_M^{\mathrm{rand}}$ represent the mean and standard deviation of $M$ in a set of equivalent random networks, respectively~\cite{Juszczyszyn2018}. In other words, the TSP shows for each sub-structure whether it occurs more or less than would be expected at random.

\textbf{Kullback-Leibler Divergence} (KL Divergence) can be used to compare distributions, such as graph kernels or the degree distribution (i.e., distribution of number of edges per node). Assuming that the characteristics of the FCMs for two agents have been summarized through distributions $P$ and $Q$ (which are statistically independent of each other), we then take the summation over all possible values of the random variable $x$ for $Q(x)$ and $P(x)$ -- that is, the probabilities assigned to each value x by the distributions $Q$ and $P$. The output of this metric is always positive and will only equal to zero if the two distributions are identical. The greater the value of KL divergence between two distributions, the greater their difference.

The last statistical test to measure distributions is the \textbf{Kolmogorov-Smirnov} (KS) statistic, which computes the maximum difference between the two distributions and determines whether the select distributions are from the same cumulative distribution~\cite{miasnikof2023statistical}. 

\begin{table}[htb]
\caption[Past measures and measures unique to our work.]{Measures in previous work or implemented for FCM comparison for the first time. V and E are the node and edge sets of the network, respectively. $N_i$ is the neighborhood of a node $i$. $R$ and $T$ are receivers and transmitters, respectively. $A$ and $B$ are centrality rankings. All measures come from \protect\cite{tchupo2022comparing} at the exception of the last three, which we propose here to compare FCMs.}
\centering
\begin{tabular}{|l|l|}
\hline
\textbf{Measure}                   & \textbf{Equation}                                                                                           \\ \hline
Graph Density                      & $D = \frac{|E|}{2\binom{|V|}{2}} = \frac{|E|}{|V|(|V|-1)}$                  \\ \hline
Graph Kernels (TSP)                & ${Z}_M=(N_M-\left\langle {N}_M^{\mathrm{rand}}\right\rangle)/\sigma_M^{\mathrm{rand}}$                \\ \hline
Number of Concepts                 & $|V|$                    \\ \hline
Clustering Coefficient             & Average of $C_i = \frac{|{e_{jk}}|}{|N_i|\cdot(|N_i|-1)}, v_j, v_k \in N_i$ over all nodes $i$                     \\ \hline
Receiver-transmitter ratio         & $\text{ratio} = \frac{R}{T}$                            \\ \hline
Weighted Jaccard similarity        & $d_{JW}(x,y) = 1- \frac{\sum_{i}^{}min(x_{i}, y_{i})}{\sum_{i}^{}max(x_{i}, y_{i})}$  \\ \hline
Centralities and cosine similarity & $\frac{A \cdot B}{\left\Vert A \right\Vert \left\Vert B \right\Vert} $  \\ \hline
Kullback-Leibler Divergence        & $D(P\|Q) = \sum _{i=1}^{k} P_{i} \log \frac {P_{i}}{Q_{i}}$              \\ \hline
Kolmogorov-Smirnov                 & $D_n = \sup_{x \in \mathbb{R}} |F_n(x) - F(x)|$                 \\ \hline
\end{tabular}
\label{tab:metricsinnovation}
\end{table}

\section{METHODS}
\label{secGED:methodsmetrics}
\subsection{First Step: Weighing Social Ties by the Expected Similarity of Behavioral Rules}
A hybrid model is initialized by creating agents with their individual FCMs, who interact through social ties. We start by identifying whether interacting agents can be merged, thus operating a local simplification. We thus only measure the \textit{similarity for each pair of interacting agents}. That is, each social tie will be weighted using one of the 11 measures (section 2.2): graph density, graph TSP, KL for node and edge weight distributions, Jaccard and KS for edge weight distributions, cosine similarity of betweenness/degree/closeness centrality, number of concepts, clustering coefficient, receiver-transmitter ratio, KL for node, and KS for edge. For example, assume that a modeler defines similarity based on the number of concepts. In Figure~\ref{fig:FCM}, agents $A$ and $B$ have $n_A=6$ and $n_B=7$ concepts respectively thus their social tie has a weight of $\frac{abs(n_A-n_B)}{(n_A+n_B)} \approx 0.07$, which indicates a small difference. If they had the same number of concepts, the difference would be 0. As the gap grows, the value tends to 1. Note that the choice of a comparison metric is an \textit{open problem}, hence we evaluate the impact of this choice experimentally in the next section.

\begin{algorithm}[H]
\caption{Evaluate the impact of simplifying a set of agents $A$ using similarity metric $S$ and community detection $D$ onto simulation outcomes. Outcome measures produced by the algorithm are shown \textcolor{blue}{in blue}.}
\label{alg:reduce-hms}
\begin{algorithmic}
\State Run the simulation several times to account for stochasticity and produce distribution of outputs $D_{original}$
\State\textit{//Identify super-agents by weighing social ties, creating communities, and finding a median agent}
\For{every pair of interacting agents $i, j \in A$}
\State Assign to the existing edge $e_{i,j}$ a weight via the similarity metric $S(FCM_i,FCM_j)$
\EndFor
\State Assign each agent to one of $c$ non-overlapping balanced clusters from algorithm $D$ using edge weights

\For{each cluster $i = 1 \ldots c$}
\State Calculate sum of each agent's FCM concept values
\State Select as `super-agent' $a_i$ an agent with the median of these values to represent the community
\EndFor

\State\textit{//Reduce simulation model by linking super-agents, removing all other agents, and initializing super-agents}
\For{each representative agent $a_i, i=1 \ldots c$}
\State Create an edge between $a_i$ and other agents $a_j, i \neq j$ from other clusters using existing edges
\EndFor
\State \textcolor{blue}{\textbf{Measure}} the number of agents to remove $|A \setminus \{a_1, \ldots, a_c\}|$ 
\State Remove every non-representative agent $A \setminus \{a_1, \ldots, a_c\}$ 
\For{each representative agent $a_i, i=1 \ldots c$}
\State Assign the initial values of $FCM_{a_i}$
\EndFor

\State\textit{//Run the simplified simulation model and compare results with the original model}
\State Run the simulation several times to account for stochasticity and produce distribution of outputs $D_{simplified}$
\State \textcolor{blue}{\textbf{Measure}} the KL divergence between distributions $D_{simplified}$ and $D_{original}$
\State \textcolor{blue}{\textbf{Measure}} the statistical properties (mean, quartile ranges, std) of $D_{simplified}$ and $D_{original}$
\end{algorithmic}
\end{algorithm}

\subsection{Second Step: Identify Groups of Like-Minded Agents by Clustering the Weighted Social Ties}
Our goal is to reduce a group to a single representative agent. Locality is important: even if an agent had a soulmate at a given time on the other side of the world, the two should not be fused since their interactions with different people/places poses a risk to diverge over time. We posit that people who think alike and are embedded in the same context are more likely to remain similar throughout a simulation. Consequently, after assigning a weight to existing interactions between agents, we use a community detection algorithm on this network of social interactions to identify clusters of \textit{(i)} like-minded agents who \textit{(ii)} share a context. We sought \textit{balanced} clusters so that a comparable level of simplification is operated throughout the agent population, instead of merging only some massive groups into few agents (i.e., few very large communities) or operating a minimal reduction by merging pairs of agents (i.e., many small communities). In addition, we use non-overlapping clusters to ensure that each super-agent represents a distinct group of the initial agents. We considered four algorithms. The {\ttfamily chinesewhispers} is a randomized algorithm in which nodes are initially assigned different classes, then take the class that dominates in their local neighborhood~\cite{biemann2006chinese}. {\ttfamily DER} (Diffusion Entropy Reducer) applies random walks and a variant of the $k$-means algorithm until the clusters stabilize~\cite{kozdoba2015community}. {\ttfamily Paris} initially assigns each node to its own class and merges the closest classes recursively as long as the modularity metric increases (i.e., agglomerative hierarchical clustering)~\cite{bonald2018hierarchical}. Finally, {\ttfamily combo} combines the optimization strategies of other algorithms as it involves merging clusters, splitting them, or re-assigning nodes to a different cluster as long as an objective function score improves~\shortcite{sobolevsky2014general}. 

\subsection{Third Step: Aggregating Each Cluster into a Representative Super-Agent}
Each cluster represents a group of agent with similar behavioral rules and local context. We now simplify each group into a super-agent. This could be achieved by morphing the agents into a new composite agent or by identifying a representative agent. We take the latter approach by \textit{(i)} calculating the sum of the concept values for each agent's FCM and \textit{(ii)} selecting an agent in the median of these values. Once these representative agents are identified, we rebuild the simulation model by keeping only these agents. We ensure that two representative agents are connected if original agents in their communities used to interact. 

\section{RESULTS}
\label{secGED:Resultsmetrics}
\subsection{Overview}
Our objective is to assess the effect of simplification onto fidelity with respect to the original simulation results. We assume the general case of stochastic models where results consist of a \textit{distribution} of outcomes across repeated simulation runs. The impact of simplification is measured by \textit{(i)} KL Divergence between the original and new distribution, \textit{(ii)} the number of reduced nodes, \textit{(ii)} statistics of distributions (mean, quartiles, std). In addition, we plot the distribution of outcomes as a visual aid. Since modelers control two parameters of our simplification process, results are analyzed with respect to the choice of similarity measure (among 11 possibilities) and the community detection algorithm (out of four choices). Although modelers do not control the case study on which the process would be applied, \textit{characteristics} of a case study can mediate the results. In particular, our process relies on weighing existing ties between agents, hence the structure of social ties can have an effect and we investigate it through different social network structures. Algorithm 1 summarizes the generation of results and their ensuing analysis. To support replicability, our implementation of the Algorithm and its use on each case study can be found at \url{https://osf.io/hpz7c/}.

\subsection{Case Studies}
We considered two case studies that openly provide FCMs on nutrition~\cite{giabbanelli2022fastgeneration} and obesity~\cite{Giabbanelli2014}. That is, these studies provide a collection of rulesets that can make agents behave in different ways. The nutrition case study consisted of 722 unique FCMs\footnote{The authors of the CMA-ES case study defined stabilization as a change in the concept `perceived intake' of less than 0.05 between consecutive steps. The maximum number of iterations is set to 100 and a hyperbolic tangent is used as transfer function. The obesity case studied used the same transfer function and numbers, but the concept for stabilization was `obesity'.}, built from observed longitudinal data about real-world individuals using an algorithm called CMA-ES. Each FCM contains the same 15 concepts that are fully connected (Table~\ref{tab:CMAESconcepts}), but weighted differently based on each individual. In the obesity case study, an FCM was created by aggregating the perspectives of experts on physical exercise and eating behaviors (Table~\ref{tab:Obesityconcepts}). To keep the case studies comparable, we created 722 unique versions of the FCM by varying the edge weights. 

\begin{table}[htb]
\caption{The CMA-ES case study has 722 FCMs with the same fully connected concepts, but different individual weights \protect\cite{giabbanelli2022fastgeneration}.}
\centering
\begin{tabular}{|c|c|c|}
\hline
\textbf{Concept \#} & \textbf{Construct}                                                            & \textbf{Operationalization}                                                                                                                                  \\ \hline
1.                  & Awareness                                                                     & \begin{tabular}[c]{@{}c@{}}Self-awareness of number of fruits eaten\end{tabular}                                                    \\ \hline
2.                  & Attitude                                                                      & \begin{tabular}[c]{@{}c@{}}Belief that eating 2 servings of fruits daily is healthy\end{tabular}                                                   \\ \hline
3.                  & Attitude Price                                                                & \begin{tabular}[c]{@{}c@{}}Belief that eating 2 servings of fruits daily is expensive\end{tabular}                                                 \\ \hline
4.                  & \multirow{2}{*}{\begin{tabular}[c]{@{}c@{}}Self-efficacy (belief\\that in the next 6 months...)\end{tabular}}                                                & ...they can eat more fruit daily if they really want to                 \\ \cline{1-1} \cline{3-3} 
5.                  &                                                                               & ...it is difficult to eat more fruit                  \\ \hline
6.                  & \multirow{2}{*}{\begin{tabular}[c]{@{}c@{}}Social-influence (belief\\that most important people...)\end{tabular}}                                             & ...think they should eat 2 pieces of fruit daily.  \\ \cline{1-1} \cline{3-3} 
7.                  &                                                                               & ...consume two pieces of fruit per day.                   \\ \hline
8.                  & Intention                                                                     & \begin{tabular}[c]{@{}c@{}}Intention to eat two pieces of fruit per day?\end{tabular}                                                               \\ \hline
9.                  & \multirow{3}{*}{Action-planning}                                              & \begin{tabular}[c]{@{}c@{}}Clear plan for when to eat more fruit.\end{tabular}                                                    \\ \cline{1-1} \cline{3-3} 
10.                 &                                                                               & \begin{tabular}[c]{@{}c@{}}Clear plan for which fruit to eat more/less.\end{tabular}                                              \\ \cline{1-1} \cline{3-3} 
11.                 &                                                                               & \begin{tabular}[c]{@{}c@{}}Clear plan for how many fruits to eat more/less.\end{tabular}                                          \\ \hline
12.                 & \multirow{2}{*}{\begin{tabular}[c]{@{}c@{}}Coping planning (plan\\what to do when...)\end{tabular}}                                              & ...something interferes with plans to eat more fruit. \\ \cline{1-1} \cline{3-3} 
13.                 &                                                                               & ...it is difficult to eat more fruit.       \\ \hline
14.                 & \begin{tabular}[c]{@{}c@{}}Perceived availability\end{tabular} & \begin{tabular}[c]{@{}c@{}}How often are fruit products available at home?\end{tabular}                                                          \\ \hline
15.                 & Visibility at home                                                            & Visibility of fruits at home                                                                                                                                 \\ \hline
\end{tabular}
\label{tab:CMAESconcepts}
\end{table}

\begin{table*}[htbp]
\caption{Edge values for the FCM in the obesity case study \protect\cite{Giabbanelli2014}.}
\centering
\begin{tabular}{|l|l|}
\hline
\textbf{Source node}              & \textbf{List of target nodes (causal weight from -1 to 1)}       \\ \hline
Age                               & Exercise (-0.44)                                                 \\ \hline
Income                            & Exercise (0.548), Fatness perceived as negative (0.478)          \\ \hline
Fatness Perceived as Negative     & Weight discrimination (0.739)                                    \\ \hline
Belief in Personal Responsibility & Weight discrimination (0.578)                                    \\ \hline
Obesity                           & Weight discrimination (0.84), Physical health (-0.795)           \\ \hline
Weight discrimination             & Depression (0.732)                                               \\ \hline
Exercise                          & Depression (-0.649), Obesity (-0.638), Physical health (0.860)   \\ \hline
Depression                        & Anti-depressants (0.592)                                         \\ \hline
Anti-depressants                  & Obesity (0.528), Food intake (0.526)                             \\ \hline
Food intake                       & Obesity (0.637)                                                  \\ \hline
Knowledge                         & Food intake (-0.5), Exercise (0.5)                               \\ \hline
Stress                            & Depression (0.54), Food intake (0.607), Physical health (-0.694) \\ \hline
\end{tabular}
\label{tab:Obesityconcepts}
\end{table*}

We created the structure of social ties by considering three network topologies: a random Erdos-Renyi graph (low clustering and normal degree distribution), a small-world Watts-Strogatz graph (high clustering with individuals forming groups), and the scale-free Barabasi-Albert (heavily skewed degree distribution with some individuals serving as hubs). In each case, individual agents were randomly assigned an FCM. In a hybrid ABM/FCM study, social ties only express \textit{who} interacts. To define \textit{how} they interact, we need to state which concepts of an individual can be observed by a peer (i.e., which parts of an FCM influence peers) and which concepts register these observations (i.e., which parts of an FCM are influenced by peers). We created a random connection, so that each agent had an equal chance to influence a particular concept of the other agent. FCM concepts were initialized between 0 and 1, and the agent with the higher value would influence its peer with a lower value. Intuitively, agents are aligning with their peers to the upside. For example, this represents how individuals acquire \textit{more} knowledge as a result of interacting with others.

\begin{table}[htb]
\caption{Properties of communities generated in the CMA-ES case study as a function of network structure, community detection algorithm ({\ttfamily chinese-whispers}, {\ttfamily der}, {\ttfamily paris}, {\ttfamily combo}), and similarity metric.}
\resizebox{\textwidth}{!}{%
\begin{tabular}{|c|c|cccccccc|}
\hline
\multirow{2}{*}{\textbf{Measure}}                                                  & \multirow{2}{*}{{\textbf{Similarity Metric}}} &  \multicolumn{4}{c|}{\textbf{Scale-free network topology}}                                                                                                                                                                    & \multicolumn{4}{c|}{\textbf{Small-world network topology}}                                                                                                                                              \\ \cline{3-10} 
                                                                                         &                                        & \multicolumn{1}{c|}{\textbf{\begin{tabular}[c]{@{}c@{}}chinese-\\ whispers\end{tabular}}} & \multicolumn{1}{c|}{\textbf{der}} & \multicolumn{1}{c|}{\textbf{{\ttfamily paris}}} & \multicolumn{1}{c|}{\textbf{{\ttfamily combo}}} & \multicolumn{1}{c|}{\textbf{\begin{tabular}[c]{@{}c@{}}chinese-\\ whispers\end{tabular}}} & \multicolumn{1}{c|}{\textbf{der}} & \multicolumn{1}{c|}{\textbf{{\ttfamily paris}}} & \textbf{{\ttfamily combo}} \\ \hline
\multirow{5}{*}{\textbf{\begin{tabular}[c]{@{}c@{}}Average \#\\ of Agents\end{tabular}}} & \textbf{Clustering Coefficient}        & \multicolumn{1}{c|}{15.1}                                                                 & \multicolumn{1}{c|}{361.0}        & \multicolumn{1}{c|}{128.7}          & \multicolumn{1}{c|}{50.7}             & \multicolumn{1}{c|}{5.8}                                                                  & \multicolumn{1}{c|}{361.0}        & \multicolumn{1}{c|}{193.1}          & 38.5             \\ \cline{2-10} 
                                                                                         & \textbf{\# of Concepts}                & \multicolumn{1}{c|}{102.2}                                                                & \multicolumn{1}{c|}{361.0}        & \multicolumn{1}{c|}{135.6}          & \multicolumn{1}{c|}{48.9}             & \multicolumn{1}{c|}{12.1}                                                                 & \multicolumn{1}{c|}{361.0}        & \multicolumn{1}{c|}{240.7}          & 45.1             \\ \cline{2-10} 
                                                                                         & \textbf{Graph Density}                 & \multicolumn{1}{c|}{16.4}                                                                 & \multicolumn{1}{c|}{361.0}        & \multicolumn{1}{c|}{83.5}           & \multicolumn{1}{c|}{48.2}             & \multicolumn{1}{c|}{5.9}                                                                  & \multicolumn{1}{c|}{361.0}        & \multicolumn{1}{c|}{185.3}          & 36.3             \\ \cline{2-10} 
                                                                                         & \textbf{R/T Ratio}                     & \multicolumn{1}{c|}{104.7}                                                                & \multicolumn{1}{c|}{361.0}        & \multicolumn{1}{c|}{135.6}          & \multicolumn{1}{c|}{48.9}             & \multicolumn{1}{c|}{12.1}                                                                 & \multicolumn{1}{c|}{361.0}        & \multicolumn{1}{c|}{240.7}          & 45.1             \\ \cline{2-10} 
                                                                                         & \textbf{TSP}                           & \multicolumn{1}{c|}{15.7}                                                                 & \multicolumn{1}{c|}{361.0}        & \multicolumn{1}{c|}{123.2}          & \multicolumn{1}{c|}{48.1}             & \multicolumn{1}{c|}{5.9}                                                                  & \multicolumn{1}{c|}{361.0}        & \multicolumn{1}{c|}{187.7}          & 32.0             \\ \hline
\multirow{5}{*}{\textbf{\begin{tabular}[c]{@{}c@{}}Max \#\\ of Agents\end{tabular}}}     & \textbf{Clustering Coefficient}        & \multicolumn{1}{c|}{306.0}                                                                & \multicolumn{1}{c|}{414.3}        & \multicolumn{1}{c|}{268.6}          & \multicolumn{1}{c|}{123.5}            & \multicolumn{1}{c|}{13.0}                                                                 & \multicolumn{1}{c|}{375.4}        & \multicolumn{1}{c|}{235.4}          & 44.0             \\ \cline{2-10} 
                                                                                         & \textbf{\# of Concepts}                & \multicolumn{1}{c|}{532.7}                                                                & \multicolumn{1}{c|}{448.9}        & \multicolumn{1}{c|}{276.1}          & \multicolumn{1}{c|}{123.3}            & \multicolumn{1}{c|}{27.7}                                                                 & \multicolumn{1}{c|}{374.1}        & \multicolumn{1}{c|}{256.0}          & 46.0             \\ \cline{2-10} 
                                                                                         & \textbf{Graph Density}                 & \multicolumn{1}{c|}{331.6}                                                                & \multicolumn{1}{c|}{411.9}        & \multicolumn{1}{c|}{194.1}          & \multicolumn{1}{c|}{125.8}            & \multicolumn{1}{c|}{13.7}                                                                 & \multicolumn{1}{c|}{371.0}        & \multicolumn{1}{c|}{223.8}          & 43.3             \\ \cline{2-10} 
                                                                                         & \textbf{R/T Ratio}                     & \multicolumn{1}{c|}{530.0}                                                                & \multicolumn{1}{c|}{422.8}        & \multicolumn{1}{c|}{276.1}          & \multicolumn{1}{c|}{123.3}            & \multicolumn{1}{c|}{27.3}                                                                 & \multicolumn{1}{c|}{373.0}        & \multicolumn{1}{c|}{256.0}          & 46.0             \\ \cline{2-10} 
                                                                                         & \textbf{TSP}                           & \multicolumn{1}{c|}{305.8}                                                                & \multicolumn{1}{c|}{403.3}        & \multicolumn{1}{c|}{249.4}          & \multicolumn{1}{c|}{126.3}            & \multicolumn{1}{c|}{13.1}                                                                 & \multicolumn{1}{c|}{370.0}        & \multicolumn{1}{c|}{223.6}          & 41.6             \\ \hline
\multirow{5}{*}{\textbf{\begin{tabular}[c]{@{}c@{}}Min \# \\ of Agents\end{tabular}}}    & \textbf{Clustering Coefficient}        & \multicolumn{1}{c|}{1.9}                                                                  & \multicolumn{1}{c|}{307.7}        & \multicolumn{1}{c|}{22.4}           & \multicolumn{1}{c|}{11.2}             & \multicolumn{1}{c|}{3.0}                                                                  & \multicolumn{1}{c|}{346.6}        & \multicolumn{1}{c|}{150.6}          & 33.3             \\ \cline{2-10} 
                                                                                         & \textbf{\# of Concepts}                & \multicolumn{1}{c|}{2.9}                                                                  & \multicolumn{1}{c|}{273.1}        & \multicolumn{1}{c|}{30.9}           & \multicolumn{1}{c|}{6.7}              & \multicolumn{1}{c|}{2.6}                                                                  & \multicolumn{1}{c|}{347.9}        & \multicolumn{1}{c|}{210.0}          & 45.0             \\ \cline{2-10} 
                                                                                         & \textbf{Graph Density}                 & \multicolumn{1}{c|}{1.9}                                                                  & \multicolumn{1}{c|}{310.1}        & \multicolumn{1}{c|}{11.6}           & \multicolumn{1}{c|}{11.9}             & \multicolumn{1}{c|}{3.0}                                                                  & \multicolumn{1}{c|}{351.0}        & \multicolumn{1}{c|}{153.8}          & 29.6             \\ \cline{2-10} 
                                                                                         & \textbf{R/T Ratio}                     & \multicolumn{1}{c|}{2.8}                                                                  & \multicolumn{1}{c|}{299.2}        & \multicolumn{1}{c|}{30.9}           & \multicolumn{1}{c|}{6.7}              & \multicolumn{1}{c|}{2.5}                                                                  & \multicolumn{1}{c|}{349.0}        & \multicolumn{1}{c|}{210.0}          & 45.0             \\ \cline{2-10} 
                                                                                         & \textbf{TSP}                           & \multicolumn{1}{c|}{1.9}                                                                  & \multicolumn{1}{c|}{318.7}        & \multicolumn{1}{c|}{17.9}           & \multicolumn{1}{c|}{7.9}              & \multicolumn{1}{c|}{3.0}                                                                  & \multicolumn{1}{c|}{352.0}        & \multicolumn{1}{c|}{147.5}          & 21.6             \\ \hline
\multirow{5}{*}{\textbf{\begin{tabular}[c]{@{}c@{}}\# of\\ Communities\end{tabular}}}    & \textbf{Clustering Coefficient}        & \multicolumn{1}{c|}{55.2}                                                                 & \multicolumn{1}{c|}{2.0}          & \multicolumn{1}{c|}{15.7}           & \multicolumn{1}{c|}{14.6}             & \multicolumn{1}{c|}{123.8}                                                                & \multicolumn{1}{c|}{2.0}          & \multicolumn{1}{c|}{4.3}            & 18.8             \\ \cline{2-10} 
                                                                                         & \textbf{\# of Concepts}                & \multicolumn{1}{c|}{10.6}                                                                 & \multicolumn{1}{c|}{2.0}          & \multicolumn{1}{c|}{13.1}           & \multicolumn{1}{c|}{15.1}             & \multicolumn{1}{c|}{59.8}                                                                 & \multicolumn{1}{c|}{2.0}          & \multicolumn{1}{c|}{3.0}            & 16.0             \\ \cline{2-10} 
                                                                                         & \textbf{Graph Density}                 & \multicolumn{1}{c|}{51.4}                                                                 & \multicolumn{1}{c|}{2.0}          & \multicolumn{1}{c|}{19.9}           & \multicolumn{1}{c|}{15.3}             & \multicolumn{1}{c|}{122.9}                                                                & \multicolumn{1}{c|}{2.0}          & \multicolumn{1}{c|}{4.2}            & 19.9             \\ \cline{2-10} 
                                                                                         & \textbf{R/T Ratio}                     & \multicolumn{1}{c|}{10.8}                                                                 & \multicolumn{1}{c|}{2.0}          & \multicolumn{1}{c|}{13.1}           & \multicolumn{1}{c|}{15.1}             & \multicolumn{1}{c|}{59.9}                                                                 & \multicolumn{1}{c|}{2.0}          & \multicolumn{1}{c|}{3.0}            & 16.0             \\ \cline{2-10} 
                                                                                         & \textbf{TSP}                           & \multicolumn{1}{c|}{54.3}                                                                 & \multicolumn{1}{c|}{2.0}          & \multicolumn{1}{c|}{10.0}           & \multicolumn{1}{c|}{15.3}             & \multicolumn{1}{c|}{123.1}                                                                & \multicolumn{1}{c|}{2.0}          & \multicolumn{1}{c|}{4.0}            & 22.6             \\ \hline
\end{tabular}%
}
\label{tab:comm_nums}
\end{table}

\subsection{Joint Effect of Community Detection Algorithm, Similarity Metric, and Network Structure}
The communities depend on the structure of social ties, the choice of a measure that assigns similarity weights to these ties, and the choice of algorithm to detect communities from weighted social ties. We measure the interplay of these effects on the CMA-ES case study, as shown in Table~\ref{tab:comm_nums} for two network structures. Communities are characterized by the average/min/max number of agents in each community and the number of communities. The results show that {\ttfamily DER} is not usable as it always yields only two communities. We also note that {\ttfamily combo} does not respond to the choice of the similarity metric, which is undesirable since the notion of a like-minded group should be largely driven by the definition of similarity. For this reason, we identify {\ttfamily chinesewhispers} as the most promising option and use it in our experiments.

\subsection{Fidelity of Simulation Results Between the Simplified and the Original Model}
\begin{figure}[ht!]
    \centering
    \includegraphics[width= \textwidth]{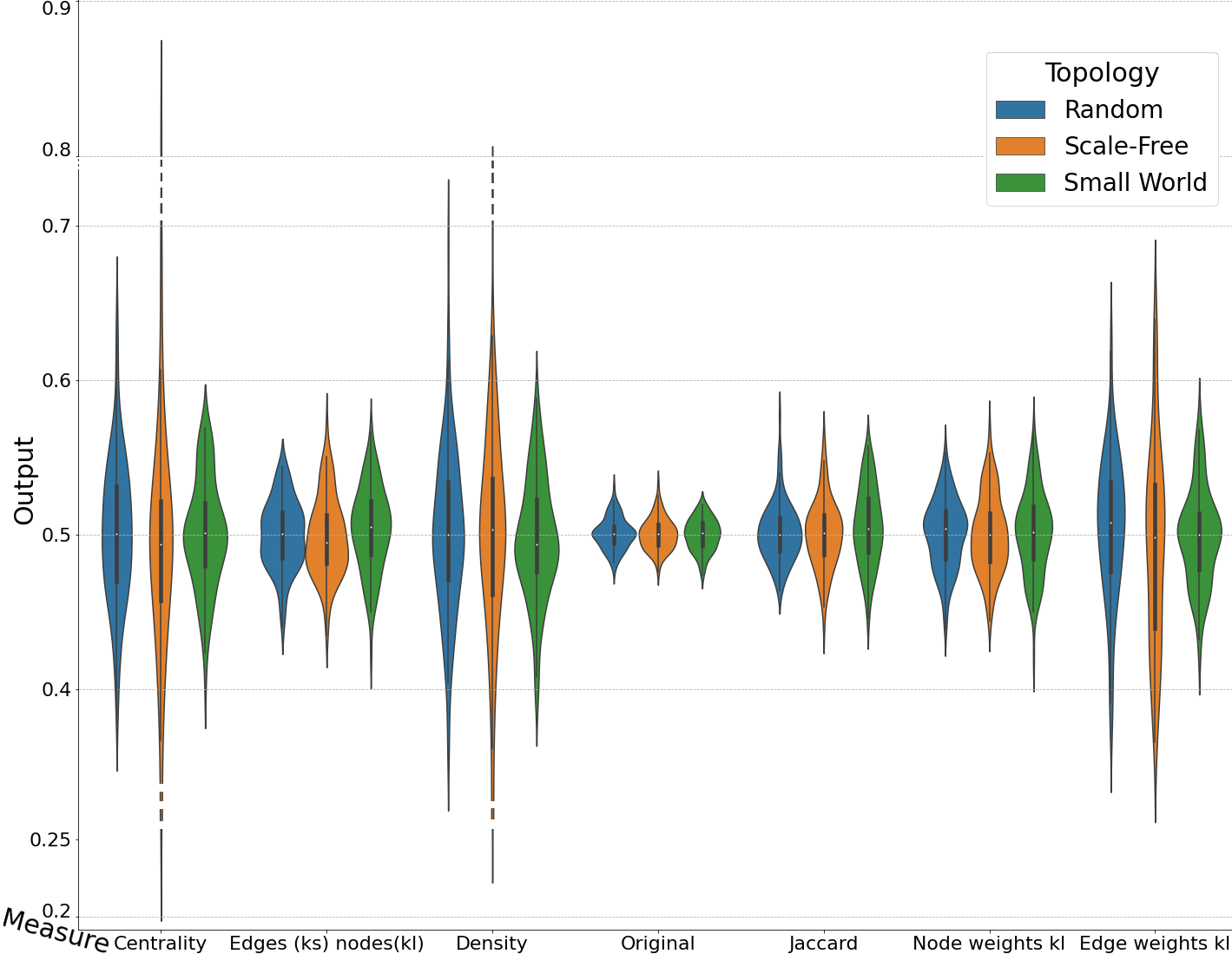}
    \caption{In the obesity case study, mean simulation outcomes in the original model are comparable with the simplified model. However, simplified models have more uncertainty, as shown by their wider distributions.}
    \label{fig:Obesity_violin}
\end{figure}

We computed the distribution of outputs from the original and simplified models based on 100 runs. In both case studies, we observed that means are similar regardless of the case study. Based on the KL divergence, the distributions are similar (Table~\ref{tab:KLcomparison}). We further analyzed these results by plotting the most similar distributions. We found that the main effect of the simplification is a \textit{greater uncertainty around the mean} compared to the original model, as illustrated in Figure~\ref{fig:Obesity_violin} for the obesity case study. Complete results on the distributions are provided in Table~\ref{tab:heavystats}. The extent of the uncertainty depends on the choice of similarity measure, as some yield excessive levels of uncertainty (e.g., centrality and cosine similarity, density) while others have more contained increases. Interestingly, we note that the KL divergence between the distribution of \textit{node weights} had about the same uncertainty as the Jaccard for the distribution of \textit{edge weights}. This means that the causal mechanisms used by an agent (edge weights) were about as useful as the initial traits of an agent (node weights) to determine the similarity of their simulated behaviors. 

\begin{table}[htb]
\caption{Entropy of the KL divergence of the distribution of outputs in the simplified vs. original model. Note that the entropy is expressed on a scale of $10^{-4}$, hence a value of `9' should be interpreted as 0.0009.}
\centering
\begin{tabular}{|ll|l|l|l|l|l|l|l|l|l|l|l|}
\hline
\multicolumn{2}{|l|}{}                                         & \rotatebox{62}{KS Edges} & \rotatebox{62}{Jaccard} & \rotatebox{62}{Centrality} & \rotatebox{62}{Density} & \rotatebox{62}{Edge KL} & \rotatebox{62}{TSP} & \rotatebox{62}{Node KL} & \rotatebox{62}{Graphs} & \rotatebox{62}{RT}  & \rotatebox{62}{Clustering} & \rotatebox{62}{\#nodes} \\ \hline
\multicolumn{1}{|l|}{\multirow{2}{*}{\rotatebox{90}{SF}}}  & CMAES     & 9        & 10      & 8          & 9       & 11      & 7   & 10      & 13     & 158 & 10         & 90      \\ \cline{2-13} 
\multicolumn{1}{|l|}{}                             & Nutrition & 82       & 8       & 151        & 107     & 88      & 85  & 9       & 10     & 171 & 129        & 125     \\ \hline
\multicolumn{1}{|l|}{\multirow{2}{*}{\rotatebox{90}{SM}}} & CMAES     & 9        & 11      & 9          & 11      & 9       & 10  & 10      & 7      & 23  & 12         & 24      \\ \cline{2-13} 
\multicolumn{1}{|l|}{}                             & Nutrition & 29       & 9       & 25         & 28      & 18      & 24  & 10      & 10     & 24  & 21         & 25      \\ \hline
\multicolumn{1}{|l|}{\multirow{2}{*}{\rotatebox{90}{Rand}}}      & CMAES     & 9        & 12      & 6          & 6       & 8       & 6   & 6       & 9      & 47  & 5          & 36      \\ \cline{2-13} 
\multicolumn{1}{|l|}{}                             & Nutrition & 48       & 7       & 41         & 50      & 44      & 36  & 9       & 9      & 42  & 46         & 32      \\ \hline
\end{tabular}
\label{tab:KLcomparison}
\end{table}

\begin{table}[htb]
\caption[Final simulation output based on 100 repeats.]{Characteristics of the distributions of simulation outputs in the simplified model}
\resizebox{\textwidth}{!}{%
\begin{tabular}{|c|c|ccccccc|c|ccccccc|}
\hline
                                       &                           & \multicolumn{7}{c|}{{\textbf{Output of the nutrition case study}}}                                                                                                                                                                                            &  & \multicolumn{7}{c|}{{\textbf{Output of the CMAES case study}}}                                                                                                                                                                                                \\ \hline
                & {\ul{ \textbf{Measure}}}    & \multicolumn{1}{c|}{\textbf{mean}} & \multicolumn{1}{c|}{\textbf{std}} & \multicolumn{1}{c|}{\textbf{min}} & \multicolumn{1}{c|}{\textbf{25\%}} & \multicolumn{1}{c|}{\textbf{50\%}} & \multicolumn{1}{c|}{\textbf{75\%}} & \textbf{max} &  & \multicolumn{1}{c|}{\textbf{mean}} & \multicolumn{1}{c|}{\textbf{std}} & \multicolumn{1}{c|}{\textbf{min}} & \multicolumn{1}{c|}{\textbf{25\%}} & \multicolumn{1}{c|}{\textbf{50\%}} & \multicolumn{1}{c|}{\textbf{75\%}} & \textbf{max} \\ \hline
\multirow{12}{*}{\rotatebox{90}{\textbf{Random}}}      & \textbf{Original}         & \multicolumn{1}{c|}{.501}         & \multicolumn{1}{c|}{.010}        & \multicolumn{1}{c|}{.476}        & \multicolumn{1}{c|}{.494}         & \multicolumn{1}{c|}{.501}         & \multicolumn{1}{c|}{.506}         & .531        &  & \multicolumn{1}{c|}{.502}         & \multicolumn{1}{c|}{.010}        & \multicolumn{1}{c|}{.479}        & \multicolumn{1}{c|}{.495}         & \multicolumn{1}{c|}{.502}         & \multicolumn{1}{c|}{.508}         & .529        \\ \cline{2-17} 
                                       & \textbf{centrality}       & \multicolumn{1}{c|}{.502}         & \multicolumn{1}{c|}{.047}        & \multicolumn{1}{c|}{.385}        & \multicolumn{1}{c|}{.470}         & \multicolumn{1}{c|}{.501}         & \multicolumn{1}{c|}{.531}         & .643        &  & \multicolumn{1}{c|}{.504}         & \multicolumn{1}{c|}{.020}        & \multicolumn{1}{c|}{.444}        & \multicolumn{1}{c|}{.494}         & \multicolumn{1}{c|}{.504}         & \multicolumn{1}{c|}{.516}         & .581        \\ \cline{2-17} 
                                       & \textbf{clustering}       & \multicolumn{1}{c|}{.503}         & \multicolumn{1}{c|}{.048}        & \multicolumn{1}{c|}{.382}        & \multicolumn{1}{c|}{.474}         & \multicolumn{1}{c|}{.505}         & \multicolumn{1}{c|}{.529}         & .623        &  & \multicolumn{1}{c|}{.502}         & \multicolumn{1}{c|}{.019}        & \multicolumn{1}{c|}{.450}        & \multicolumn{1}{c|}{.490}         & \multicolumn{1}{c|}{.498}         & \multicolumn{1}{c|}{.514}         & .556        \\ \cline{2-17} 
                                       & \textbf{compare graphs}   & \multicolumn{1}{c|}{.500}         & \multicolumn{1}{c|}{.022}        & \multicolumn{1}{c|}{.440}        & \multicolumn{1}{c|}{.485}         & \multicolumn{1}{c|}{.500}         & \multicolumn{1}{c|}{.515}         & .544        &  & \multicolumn{1}{c|}{.502}         & \multicolumn{1}{c|}{.024}        & \multicolumn{1}{c|}{.442}        & \multicolumn{1}{c|}{.488}         & \multicolumn{1}{c|}{.501}         & \multicolumn{1}{c|}{.519}         & .558        \\ \cline{2-17} 
                                       & \textbf{concepts}         & \multicolumn{1}{c|}{.502}         & \multicolumn{1}{c|}{.041}        & \multicolumn{1}{c|}{.420}        & \multicolumn{1}{c|}{.473}         & \multicolumn{1}{c|}{.507}         & \multicolumn{1}{c|}{.530}         & .588        &  & \multicolumn{1}{c|}{.499}         & \multicolumn{1}{c|}{.044}        & \multicolumn{1}{c|}{.389}        & \multicolumn{1}{c|}{.465}         & \multicolumn{1}{c|}{.496}         & \multicolumn{1}{c|}{.530}         & .604        \\ \cline{2-17} 
                                       & \textbf{density}          & \multicolumn{1}{c|}{.501}         & \multicolumn{1}{c|}{.052}        & \multicolumn{1}{c|}{.363}        & \multicolumn{1}{c|}{.471}         & \multicolumn{1}{c|}{.500}         & \multicolumn{1}{c|}{.534}         & .688        &  & \multicolumn{1}{c|}{.500}         & \multicolumn{1}{c|}{.018}        & \multicolumn{1}{c|}{.462}        & \multicolumn{1}{c|}{.487}         & \multicolumn{1}{c|}{.500}         & \multicolumn{1}{c|}{.512}         & .549        \\ \cline{2-17} 
                                       & \textbf{edge weight kl}   & \multicolumn{1}{c|}{.504}         & \multicolumn{1}{c|}{.047}        & \multicolumn{1}{c|}{.371}        & \multicolumn{1}{c|}{.476}         & \multicolumn{1}{c|}{.508}         & \multicolumn{1}{c|}{.534}         & .626        &  & \multicolumn{1}{c|}{.500}         & \multicolumn{1}{c|}{.023}        & \multicolumn{1}{c|}{.439}        & \multicolumn{1}{c|}{.485}         & \multicolumn{1}{c|}{.500}         & \multicolumn{1}{c|}{.516}         & .547        \\ \cline{2-17} 
                                       & \textbf{Jaccard} & \multicolumn{1}{c|}{.501}         & \multicolumn{1}{c|}{.019}        & \multicolumn{1}{c|}{.464}        & \multicolumn{1}{c|}{.489}         & \multicolumn{1}{c|}{.500}         & \multicolumn{1}{c|}{.511}         & .577        &  & \multicolumn{1}{c|}{.500}         & \multicolumn{1}{c|}{.027}        & \multicolumn{1}{c|}{.439}        & \multicolumn{1}{c|}{.481}         & \multicolumn{1}{c|}{.500}         & \multicolumn{1}{c|}{.515}         & .581        \\ \cline{2-17} 
                                       & \textbf{KS edges}          & \multicolumn{1}{c|}{.504}         & \multicolumn{1}{c|}{.051}        & \multicolumn{1}{c|}{.373}        & \multicolumn{1}{c|}{.472}         & \multicolumn{1}{c|}{.500}         & \multicolumn{1}{c|}{.537}         & .624        &  & \multicolumn{1}{c|}{.504}         & \multicolumn{1}{c|}{.023}        & \multicolumn{1}{c|}{.445}        & \multicolumn{1}{c|}{.488}         & \multicolumn{1}{c|}{.502}         & \multicolumn{1}{c|}{.519}         & .578        \\ \cline{2-17} 
                                       & \textbf{node weights kl}  & \multicolumn{1}{c|}{.501}         & \multicolumn{1}{c|}{.022}        & \multicolumn{1}{c|}{.439}        & \multicolumn{1}{c|}{.484}         & \multicolumn{1}{c|}{.504}         & \multicolumn{1}{c|}{.515}         & .553        &  & \multicolumn{1}{c|}{.501}         & \multicolumn{1}{c|}{.021}        & \multicolumn{1}{c|}{.439}        & \multicolumn{1}{c|}{.487}         & \multicolumn{1}{c|}{.502}         & \multicolumn{1}{c|}{.517}         & .559        \\ \cline{2-17} 
                                       & \textbf{rt ratio}         & \multicolumn{1}{c|}{.504}         & \multicolumn{1}{c|}{.046}        & \multicolumn{1}{c|}{.387}        & \multicolumn{1}{c|}{.476}         & \multicolumn{1}{c|}{.505}         & \multicolumn{1}{c|}{.529}         & .598        &  & \multicolumn{1}{c|}{.505}         & \multicolumn{1}{c|}{.051}        & \multicolumn{1}{c|}{.379}        & \multicolumn{1}{c|}{.473}         & \multicolumn{1}{c|}{.506}         & \multicolumn{1}{c|}{.537}         & .643        \\ \cline{2-17} 
                                       & \textbf{tsp}              & \multicolumn{1}{c|}{.501}         & \multicolumn{1}{c|}{.044}        & \multicolumn{1}{c|}{.408}        & \multicolumn{1}{c|}{.466}         & \multicolumn{1}{c|}{.497}         & \multicolumn{1}{c|}{.527}         & .634        &  & \multicolumn{1}{c|}{.499}         & \multicolumn{1}{c|}{.019}        & \multicolumn{1}{c|}{.456}        & \multicolumn{1}{c|}{.486}         & \multicolumn{1}{c|}{.496}         & \multicolumn{1}{c|}{.513}         & .551        \\ \hline
\multirow{12}{*}{\rotatebox{90}{\textbf{Scale-Free}}}  & \textbf{Original}         & \multicolumn{1}{c|}{.500}         & \multicolumn{1}{c|}{.010}        & \multicolumn{1}{c|}{.476}        & \multicolumn{1}{c|}{.493}         & \multicolumn{1}{c|}{.500}         & \multicolumn{1}{c|}{.507}         & .533        &  & \multicolumn{1}{c|}{.500}         & \multicolumn{1}{c|}{.010}        & \multicolumn{1}{c|}{.475}        & \multicolumn{1}{c|}{.493}         & \multicolumn{1}{c|}{.499}         & \multicolumn{1}{c|}{.507}         & .520        \\ \cline{2-17} 
                                       & \textbf{centrality}       & \multicolumn{1}{c|}{.493}         & \multicolumn{1}{c|}{.085}        & \multicolumn{1}{c|}{.265}        & \multicolumn{1}{c|}{.457}         & \multicolumn{1}{c|}{.494}         & \multicolumn{1}{c|}{.522}         & .807        &  & \multicolumn{1}{c|}{.499}         & \multicolumn{1}{c|}{.022}        & \multicolumn{1}{c|}{.448}        & \multicolumn{1}{c|}{.483}         & \multicolumn{1}{c|}{.500}         & \multicolumn{1}{c|}{.514}         & .546        \\ \cline{2-17} 
                                       & \textbf{clustering}       & \multicolumn{1}{c|}{.494}         & \multicolumn{1}{c|}{.076}        & \multicolumn{1}{c|}{.200}        & \multicolumn{1}{c|}{.461}         & \multicolumn{1}{c|}{.498}         & \multicolumn{1}{c|}{.528}         & .736        &  & \multicolumn{1}{c|}{.499}         & \multicolumn{1}{c|}{.024}        & \multicolumn{1}{c|}{.443}        & \multicolumn{1}{c|}{.484}         & \multicolumn{1}{c|}{.499}         & \multicolumn{1}{c|}{.518}         & .546        \\ \cline{2-17} 
                                       & \textbf{compare graphs}   & \multicolumn{1}{c|}{.499}         & \multicolumn{1}{c|}{.026}        & \multicolumn{1}{c|}{.435}        & \multicolumn{1}{c|}{.481}         & \multicolumn{1}{c|}{.495}         & \multicolumn{1}{c|}{.513}         & .571        &  & \multicolumn{1}{c|}{.500}         & \multicolumn{1}{c|}{.028}        & \multicolumn{1}{c|}{.437}        & \multicolumn{1}{c|}{.484}         & \multicolumn{1}{c|}{.500}         & \multicolumn{1}{c|}{.516}         & .568        \\ \cline{2-17} 
                                       & \textbf{concepts}         & \multicolumn{1}{c|}{.504}         & \multicolumn{1}{c|}{.079}        & \multicolumn{1}{c|}{.266}        & \multicolumn{1}{c|}{.475}         & \multicolumn{1}{c|}{.505}         & \multicolumn{1}{c|}{.541}         & .712        &  & \multicolumn{1}{c|}{.504}         & \multicolumn{1}{c|}{.070}        & \multicolumn{1}{c|}{.267}        & \multicolumn{1}{c|}{.482}         & \multicolumn{1}{c|}{.500}         & \multicolumn{1}{c|}{.522}         & .832        \\ \cline{2-17} 
                                       & \textbf{density}          & \multicolumn{1}{c|}{.500}         & \multicolumn{1}{c|}{.073}        & \multicolumn{1}{c|}{.280}        & \multicolumn{1}{c|}{.461}         & \multicolumn{1}{c|}{.503}         & \multicolumn{1}{c|}{.537}         & .748        &  & \multicolumn{1}{c|}{.498}         & \multicolumn{1}{c|}{.024}        & \multicolumn{1}{c|}{.425}        & \multicolumn{1}{c|}{.483}         & \multicolumn{1}{c|}{.497}         & \multicolumn{1}{c|}{.510}         & .559        \\ \cline{2-17} 
                                       & \textbf{edge weight kl}   & \multicolumn{1}{c|}{.492}         & \multicolumn{1}{c|}{.065}        & \multicolumn{1}{c|}{.365}        & \multicolumn{1}{c|}{.440}         & \multicolumn{1}{c|}{.498}         & \multicolumn{1}{c|}{.532}         & .639        &  & \multicolumn{1}{c|}{.499}         & \multicolumn{1}{c|}{.025}        & \multicolumn{1}{c|}{.416}        & \multicolumn{1}{c|}{.484}         & \multicolumn{1}{c|}{.499}         & \multicolumn{1}{c|}{.513}         & .561        \\ \cline{2-17} 
                                       & \textbf{Jaccard} & \multicolumn{1}{c|}{.500}         & \multicolumn{1}{c|}{.022}        & \multicolumn{1}{c|}{.441}        & \multicolumn{1}{c|}{.487}         & \multicolumn{1}{c|}{.501}         & \multicolumn{1}{c|}{.513}         & .562        &  & \multicolumn{1}{c|}{.498}         & \multicolumn{1}{c|}{.025}        & \multicolumn{1}{c|}{.441}        & \multicolumn{1}{c|}{.479}         & \multicolumn{1}{c|}{.500}         & \multicolumn{1}{c|}{.514}         & .564        \\ \cline{2-17} 
                                       & \textbf{KS edges}          & \multicolumn{1}{c|}{.500}         & \multicolumn{1}{c|}{.064}        & \multicolumn{1}{c|}{.350}        & \multicolumn{1}{c|}{.454}         & \multicolumn{1}{c|}{.493}         & \multicolumn{1}{c|}{.548}         & .652        &  & \multicolumn{1}{c|}{.497}         & \multicolumn{1}{c|}{.022}        & \multicolumn{1}{c|}{.448}        & \multicolumn{1}{c|}{.485}         & \multicolumn{1}{c|}{.496}         & \multicolumn{1}{c|}{.509}         & .592        \\ \cline{2-17} 
                                       & \textbf{node weights kl}  & \multicolumn{1}{c|}{.500}         & \multicolumn{1}{c|}{.025}        & \multicolumn{1}{c|}{.444}        & \multicolumn{1}{c|}{.483}         & \multicolumn{1}{c|}{.500}         & \multicolumn{1}{c|}{.514}         & .567        &  & \multicolumn{1}{c|}{.499}         & \multicolumn{1}{c|}{.025}        & \multicolumn{1}{c|}{.431}        & \multicolumn{1}{c|}{.482}         & \multicolumn{1}{c|}{.499}         & \multicolumn{1}{c|}{.518}         & .555        \\ \cline{2-17} 
                                       & \textbf{rt ratio}         & \multicolumn{1}{c|}{.496}         & \multicolumn{1}{c|}{.098}        & \multicolumn{1}{c|}{.251}        & \multicolumn{1}{c|}{.442}         & \multicolumn{1}{c|}{.493}         & \multicolumn{1}{c|}{.519}         & .895        &  & \multicolumn{1}{c|}{.504}         & \multicolumn{1}{c|}{.089}        & \multicolumn{1}{c|}{.201}        & \multicolumn{1}{c|}{.461}         & \multicolumn{1}{c|}{.494}         & \multicolumn{1}{c|}{.535}         & .903        \\ \cline{2-17} 
                                       & \textbf{tsp}              & \multicolumn{1}{c|}{.504}         & \multicolumn{1}{c|}{.068}        & \multicolumn{1}{c|}{.316}        & \multicolumn{1}{c|}{.472}         & \multicolumn{1}{c|}{.502}         & \multicolumn{1}{c|}{.532}         & .768        &  & \multicolumn{1}{c|}{.499}         & \multicolumn{1}{c|}{.022}        & \multicolumn{1}{c|}{.438}        & \multicolumn{1}{c|}{.486}         & \multicolumn{1}{c|}{.501}         & \multicolumn{1}{c|}{.514}         & .550        \\ \hline
\multirow{12}{*}{\rotatebox{90}{\textbf{Small-world}}} & \textbf{Original}         & \multicolumn{1}{c|}{.500}         & \multicolumn{1}{c|}{.010}        & \multicolumn{1}{c|}{.473}        & \multicolumn{1}{c|}{.492}         & \multicolumn{1}{c|}{.501}         & \multicolumn{1}{c|}{.508}         & .520        &  & \multicolumn{1}{c|}{.502}         & \multicolumn{1}{c|}{.011}        & \multicolumn{1}{c|}{.478}        & \multicolumn{1}{c|}{.496}         & \multicolumn{1}{c|}{.502}         & \multicolumn{1}{c|}{.507}         & .535        \\ \cline{2-17} 
                                       & \textbf{centrality}       & \multicolumn{1}{c|}{.500}         & \multicolumn{1}{c|}{.035}        & \multicolumn{1}{c|}{.403}        & \multicolumn{1}{c|}{.480}         & \multicolumn{1}{c|}{.501}         & \multicolumn{1}{c|}{.521}         & .569        &  & \multicolumn{1}{c|}{.498}         & \multicolumn{1}{c|}{.023}        & \multicolumn{1}{c|}{.447}        & \multicolumn{1}{c|}{.482}         & \multicolumn{1}{c|}{.498}         & \multicolumn{1}{c|}{.514}         & .554        \\ \cline{2-17} 
                                       & \textbf{clustering}       & \multicolumn{1}{c|}{.499}         & \multicolumn{1}{c|}{.035}        & \multicolumn{1}{c|}{.426}        & \multicolumn{1}{c|}{.472}         & \multicolumn{1}{c|}{.496}         & \multicolumn{1}{c|}{.522}         & .597        &  & \multicolumn{1}{c|}{.501}         & \multicolumn{1}{c|}{.025}        & \multicolumn{1}{c|}{.445}        & \multicolumn{1}{c|}{.483}         & \multicolumn{1}{c|}{.503}         & \multicolumn{1}{c|}{.520}         & .555        \\ \cline{2-17} 
                                       & \textbf{compare graphs}   & \multicolumn{1}{c|}{.504}         & \multicolumn{1}{c|}{.027}        & \multicolumn{1}{c|}{.422}        & \multicolumn{1}{c|}{.487}         & \multicolumn{1}{c|}{.505}         & \multicolumn{1}{c|}{.522}         & .566        &  & \multicolumn{1}{c|}{.504}         & \multicolumn{1}{c|}{.022}        & \multicolumn{1}{c|}{.433}        & \multicolumn{1}{c|}{.488}         & \multicolumn{1}{c|}{.506}         & \multicolumn{1}{c|}{.518}         & .560        \\ \cline{2-17} 
                                       & \textbf{concepts}         & \multicolumn{1}{c|}{.502}         & \multicolumn{1}{c|}{.039}        & \multicolumn{1}{c|}{.416}        & \multicolumn{1}{c|}{.479}         & \multicolumn{1}{c|}{.498}         & \multicolumn{1}{c|}{.530}         & .592        &  & \multicolumn{1}{c|}{.497}         & \multicolumn{1}{c|}{.036}        & \multicolumn{1}{c|}{.430}        & \multicolumn{1}{c|}{.471}         & \multicolumn{1}{c|}{.497}         & \multicolumn{1}{c|}{.516}         & .635        \\ \cline{2-17} 
                                       & \textbf{density}          & \multicolumn{1}{c|}{.497}         & \multicolumn{1}{c|}{.039}        & \multicolumn{1}{c|}{.394}        & \multicolumn{1}{c|}{.476}         & \multicolumn{1}{c|}{.494}         & \multicolumn{1}{c|}{.523}         & .588        &  & \multicolumn{1}{c|}{.501}         & \multicolumn{1}{c|}{.024}        & \multicolumn{1}{c|}{.444}        & \multicolumn{1}{c|}{.484}         & \multicolumn{1}{c|}{.501}         & \multicolumn{1}{c|}{.517}         & .573        \\ \cline{2-17} 
                                       & \textbf{edge weight kl}   & \multicolumn{1}{c|}{.499}         & \multicolumn{1}{c|}{.031}        & \multicolumn{1}{c|}{.421}        & \multicolumn{1}{c|}{.478}         & \multicolumn{1}{c|}{.500}         & \multicolumn{1}{c|}{.514}         & .577        &  & \multicolumn{1}{c|}{.500}         & \multicolumn{1}{c|}{.023}        & \multicolumn{1}{c|}{.447}        & \multicolumn{1}{c|}{.483}         & \multicolumn{1}{c|}{.502}         & \multicolumn{1}{c|}{.515}         & .564        \\ \cline{2-17} 
                                       & \textbf{Jaccard} & \multicolumn{1}{c|}{.505}         & \multicolumn{1}{c|}{.025}        & \multicolumn{1}{c|}{.446}        & \multicolumn{1}{c|}{.488}         & \multicolumn{1}{c|}{.504}         & \multicolumn{1}{c|}{.523}         & .557        &  & \multicolumn{1}{c|}{.504}         & \multicolumn{1}{c|}{.028}        & \multicolumn{1}{c|}{.452}        & \multicolumn{1}{c|}{.485}         & \multicolumn{1}{c|}{.498}         & \multicolumn{1}{c|}{.526}         & .586        \\ \cline{2-17} 
                                       & \textbf{KS edges}          & \multicolumn{1}{c|}{.501}         & \multicolumn{1}{c|}{.039}        & \multicolumn{1}{c|}{.421}        & \multicolumn{1}{c|}{.473}         & \multicolumn{1}{c|}{.502}         & \multicolumn{1}{c|}{.525}         & .612        &  & \multicolumn{1}{c|}{.501}         & \multicolumn{1}{c|}{.023}        & \multicolumn{1}{c|}{.444}        & \multicolumn{1}{c|}{.486}         & \multicolumn{1}{c|}{.502}         & \multicolumn{1}{c|}{.514}         & .554        \\ \cline{2-17} 
                                       & \textbf{node weights kl}  & \multicolumn{1}{c|}{.501}         & \multicolumn{1}{c|}{.027}        & \multicolumn{1}{c|}{.420}        & \multicolumn{1}{c|}{.484}         & \multicolumn{1}{c|}{.502}         & \multicolumn{1}{c|}{.518}         & .568        &  & \multicolumn{1}{c|}{.502}         & \multicolumn{1}{c|}{.025}        & \multicolumn{1}{c|}{.455}        & \multicolumn{1}{c|}{.483}         & \multicolumn{1}{c|}{.501}         & \multicolumn{1}{c|}{.518}         & .572        \\ \cline{2-17} 
                                       & \textbf{rt ratio}         & \multicolumn{1}{c|}{.501}         & \multicolumn{1}{c|}{.037}        & \multicolumn{1}{c|}{.389}        & \multicolumn{1}{c|}{.477}         & \multicolumn{1}{c|}{.500}         & \multicolumn{1}{c|}{.528}         & .593        &  & \multicolumn{1}{c|}{.503}         & \multicolumn{1}{c|}{.037}        & \multicolumn{1}{c|}{.383}        & \multicolumn{1}{c|}{.479}         & \multicolumn{1}{c|}{.504}         & \multicolumn{1}{c|}{.532}         & .575        \\ \cline{2-17} 
                                       & \textbf{tsp}              & \multicolumn{1}{c|}{.499}         & \multicolumn{1}{c|}{.036}        & \multicolumn{1}{c|}{.418}        & \multicolumn{1}{c|}{.476}         & \multicolumn{1}{c|}{.495}         & \multicolumn{1}{c|}{.527}         & .583        &  & \multicolumn{1}{c|}{.502}         & \multicolumn{1}{c|}{.023}        & \multicolumn{1}{c|}{.450}        & \multicolumn{1}{c|}{.484}         & \multicolumn{1}{c|}{.503}         & \multicolumn{1}{c|}{.517}         & .561        \\ \hline
\end{tabular}%
}
\label{tab:heavystats}
\end{table}

\section{Discussion and Conclusion}

Scaling a simulation by creating a small number of representative super-agents that represent a collective ``has implications for model forecasts [but] is an underdeveloped field of study''~\cite{wise2023scale}. As noted by Wise \textit{et al.}, population sizes can vary hence the extent of scaling is variable. Furthermore, the dynamics of the simplified model can also differ from the original one. In our work, we considered the structure of social ties, the definition of a group (i.e., the clustering algorithm) and the definition of similarity. We showed that {\ttfamily chinese-whispers} can create groups of varying sizes depending on the similarity and social ties. Our work complement Wise \textit{et al.} who focused on the geographical dynamics of a model, while we examined the impact of social ties in a hybrid model. We also contribute to the work of~\citeN{toupance2023system}, since their reduced model was obtained by a fixed grouping of nearby cells whereas we adapt groups based on social ties and similarity. In our work, the average simulation output was preserved and distributions of simulated outputs were similar between the simplified and original model, although the simplification increases the uncertainty. Future work should comprehensively examine the costs associated with mitigation scenarios for the uncertainty, as these costs may partly offset the benefits of our proposed simplification. For example, if a simplified simulation requires \textit{more runs} than the original model to reach the same desired confidence interval, then the performance improvements from the simplification are reduced.

We analyzed the impact of reducing the number of nodes in a hybrid ABM/FCM model based on the two parameters of our method and one mediating characteristic of case studies. We incorporated three new metrics along with existing metrics to measure the similarities of FCMs of various agents and studied the impact of clustering agents into representative super agents. Both aspects can be elaborated upon in future works. By applying our methods to more cases, modelers may start to discern additional mediating characteristics, particularly with respect to the rulesets of the agents. While the Jaccard and KL metrics were strong performers in our study (by having less uncertainty than alternatives), metrics to compare agents can continue to be developed. For instance, instead of measuring the total number of concepts, we can use a distribution approach by measuring the Hamming distance between the presence or absence of specific concepts. This could be further weighted by concept categories: an agent motivated by price and perceived availability differs from an agent acting based on social-influence, intention, and visibility.

The hybrid ABM/FCM framework is an asset for our study as the agents' behaviors are defined through simulation models that allow to leverage the rich literature on network comparison. Dozens of studies have employed the ABM/FCM paradigm~\cite{davis2019intersection}, with new case studies appearing regularly~\cite{kocabey2023hybrid}. However, there are many ways to operationalize the internal cognitive processes of an agent. ABMs can integrate causal relationships, but the representation of these relationships is not necessarily performed through a transparent network within each agent~\shortcite{antosz2022sensemaking}. Our findings may thus only apply to frameworks in which cognitive processes of the agents are \textit{explicitly} represented as a \textit{network}. For example, hybrid ABM / System Dynamics (SD) models provide a closely related approach by using a network-based simulation model to guide the decision-making activities within each agent~\cite{schieritz2003emergent,swinerd2014simulating}. Our metrics can thus apply to the ABM/SD approach, which could broaden the benefits of our approach for the simulation community. 




\end{document}